\documentclass[10pt,twocolumn,letterpaper]{article}

\usepackage{cvpr}
\usepackage{times}
\usepackage{epsfig}
\usepackage{graphicx}
\usepackage{amsmath}
\usepackage{amssymb,bm}
\usepackage{algorithm,algpseudocode,amsthm,mathrsfs,subcaption}

\newcommand{\red}[1]{\begin{normalsize}\noindent\color{red}{#1}\end{normalsize}}

\usepackage[breaklinks=true,bookmarks=false]{hyperref}

\cvprfinalcopy 

\def\cvprPaperID{1894} 

\def\x{\boldsymbol{x}}
\def\r{\boldsymbol{r}}
\def\w{\boldsymbol{w}}

\DeclareMathOperator*{\argmin}{arg\,min}
\DeclareMathOperator*{\argmax}{arg\,max}

\ifcvprfinal\fi
\begin{document}

\title{DeepFool: a simple and accurate method to fool deep neural networks}
\author{Seyed-Mohsen Moosavi-Dezfooli, Alhussein Fawzi, Pascal Frossard\\
\'Ecole Polytechnique F\'ed\'erale de Lausanne\\
{\tt\small \{seyed.moosavi,alhussein.fawzi,pascal.frossard\} at epfl.ch}
}

\maketitle
\thispagestyle{empty}

\begin{abstract}
State-of-the-art deep neural networks have achieved impressive results on many image classification tasks. However, these same architectures have been shown to be unstable to small, well sought, perturbations of the images. Despite the importance of this phenomenon, no effective methods have been proposed to accurately compute the robustness of state-of-the-art deep classifiers to such perturbations on large-scale datasets. In this paper, we fill this gap and propose the DeepFool algorithm to efficiently compute perturbations that fool deep networks, and thus reliably quantify the robustness of these classifiers. Extensive experimental results show that our approach outperforms recent methods in the task of computing adversarial perturbations and making classifiers more robust.\footnote{To encourage reproducible research, the code of DeepFool is made available at \url{http://github.com/lts4/deepfool}}
%
%
%
%
%
%
\end{abstract}

\section{Introduction}
\begin{figure}[ht!]
\center
\includegraphics[scale=0.8]{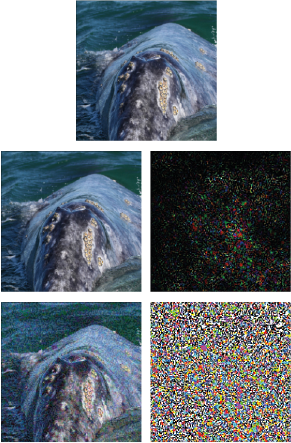}
\caption{An example of adversarial perturbations. First row: the original image $\x$ that is classified as $\hat{k}(\x)$=``whale". Second row: the image $\x+\r$ classified as $\hat{k}(\x+\r)$=``turtle" and the corresponding perturbation $\r$ computed by DeepFool. Third row: the image classified as ``turtle" and the corresponding perturbation computed by the fast gradient sign method  \cite{goodfellow2014}. DeepFool leads to a smaller perturbation.}
\label{fig:perturbed_smpl}
\end{figure}

Deep neural networks are powerful learning models that achieve state-of-the-art pattern recognition performance  in many research areas such as bioinformatics \cite{bio1,bio2}, speech \cite{sp1,sp2}, and computer vision \cite{cv1,cv2}. Though deep networks have exhibited very good performance in classification tasks, they have recently been shown to be particularly unstable to \textit{adversarial} perturbations of the data \cite{szegedy2013}. In fact, very small and often imperceptible perturbations of the data samples are sufficient to fool state-of-the-art classifiers and result in incorrect classification. (e.g., Figure \ref{fig:perturbed_smpl}). Formally, for a given classifier, we define an adversarial perturbation as the \textit{minimal} perturbation $\r$ that is sufficient to change the estimated label $\hat{k}(\x)$:
\begin{align}
\Delta(\x;\hat{k}):=\min_{\r} \| \r \|_2 \text{ subject to } \hat{k}(\x+\r) \neq \hat{k}(\x),
\label{eq:opt_robustness}
\end{align}
where $\x$ is an image and $\hat{k}(\x)$ is the estimated label. We call $\Delta(\x;\hat{k})$ the robustness of $\hat{k}$ at point $\x$. 
The robustness of classifier $\hat{k}$ is then defined as

\begin{equation}
\rho_{\text{adv}}(\hat{k})=\mathbb{E}_{\x}\frac{\Delta(\x;\hat{k})}{\|\x\|_2},
\end{equation}
where $\mathbb{E}_{\x}$ is the expectation over the distribution of data.
The study of adversarial perturbations helps us understand what features are used by a classifier.
The existence of such examples is seemingly in contradiction with the generalization ability of the learning algorithms. While deep networks achieve state-of-the-art performance in image classification tasks, they are not robust at all to small adversarial perturbations and tend to misclassify minimally perturbed data that looks visually similar to clean samples.
Though adversarial attacks are specific to the classifier, it seems that the adversarial perturbations are generalizable across different models \cite{szegedy2013}. This can actually become a real concern from a security point of view.

An accurate method for finding the adversarial perturbations is thus necessary to study and compare the robustness of different classifiers to adversarial perturbations. It might be the key to a better understanding of the limits of current architectures and to design methods to increase robustness. Despite the importance of the vulnerability of state-of-the-art classifiers to adversarial instability, no well-founded method has been proposed to compute adversarial perturbations and we fill this gap in this paper.

\vspace{2mm}

\noindent \textbf{Our main contributions are the following:}
\begin{itemize}
\item We propose a simple yet accurate method for computing and comparing the robustness of different classifiers to adversarial perturbations.
\item We perform an extensive experimental comparison, and show that 1) our method computes adversarial perturbations more reliably and efficiently than existing methods 2) augmenting training data with adversarial examples significantly increases the robustness to adversarial perturbations.
\item We show that using imprecise approaches for the computation of adversarial perturbations could lead to \textit{different and sometimes misleading conclusions} about the robustness. Hence, our method provides a better understanding of this intriguing phenomenon and of its influence factors.
\end{itemize}
We now review some of the relevant work. The phenomenon of adversarial instability was first introduced and studied in \cite{szegedy2013}. The authors estimated adversarial examples by solving penalized optimization problems and presented an analysis showing that the high complexity of neural networks might be a reason explaining the presence of adversarial examples. Unfortunately, the optimization method employed in \cite{szegedy2013} is time-consuming and therefore does not scale to large datasets. In \cite{pepik2015}, the authors showed that convolutional networks are not invariant to some sort of transformations based on the experiments done on Pascal3D+ annotations. Recently, Tsai et al. \cite{ostrich2015} provided a software to misclassify a given image in a specified class, without necessarily finding the smallest perturbation. Nguyen et al. \cite{nguyen2015} generated synthetic unrecognizable images, which are classified with high confidence. The authors of \cite{fawzi2015b} also studied a related problem of finding the minimal \textit{geometric} transformation that fools image classifiers, and provided quantitative measure of the robustness of classifiers to geometric transformations. 
Closer to our work, the authors of \cite{goodfellow2014} introduced the ``fast gradient sign" method, which computes the adversarial perturbations for a given classifier very efficiently. Despite its efficiency, this method provides only a coarse approximation of the optimal perturbation vectors. In fact, it performs a unique gradient step, which often leads to sub-optimal solutions. 
Then in an attempt to build more robust classifiers to adversarial perturbations, \cite{gu2015} introduced a smoothness penalty in the training procedure that allows to boost the robustness of the classifier. Notably, the method in \cite{szegedy2013} was applied in order to generate adversarial perturbations. We should finally mention that the phenomenon of adversarial instability also led to theoretical work in \cite{fawzi2015a} that studied the problem of adversarial perturbations on some families of classifiers, and provided upper bounds on the robustness of these classifiers. 
A deeper understanding of the phenomenon of adversarial instability for more complex classifiers is however needed; the method proposed in this work can be seen as a baseline to efficiently and accurately generate adversarial perturbations in order to better understand this phenomenon.

The rest of paper is organized as follows. In Section 2, we introduce an efficient algorithm to find adversarial perturbations in a binary classifier. The extension to the multiclass problem is provided in Section 3. In Section 4, we propose extensive experiments that confirm the accuracy of our method and outline its benefits in building more robust classifiers.

\section{DeepFool for binary classifiers}
As a multiclass classifier can be viewed as aggregation of binary classifiers, we first propose the algorithm for binary classifiers.
That is, we assume here $\hat{k}(\x) = \text{sign} (f(\x))$, where $f$ is an arbitrary scalar-valued image classification function $f: \mathbb{R}^n \rightarrow \mathbb{R}$. We also denote by $\mathscr{F} \triangleq \{ \x: f(\x) = 0 \}$ the level set at zero of $f$.
We begin by analyzing the case where $f$ is an affine classifier $f(\x) = \w^T \x + b$, and then derive the general algorithm, which can be applied to any differentiable binary classifier $f$.



\begin{figure}
\center
\includegraphics[scale=0.7]{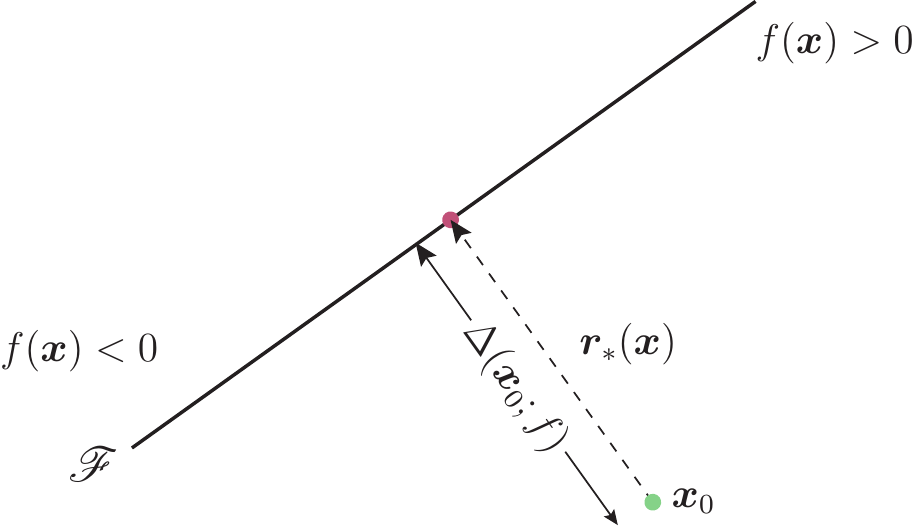}
\caption{Adversarial examples for a linear binary classifier.}
\label{fig:binary_lin}
\end{figure}

\begin{figure}
\center
\includegraphics[scale=0.5]{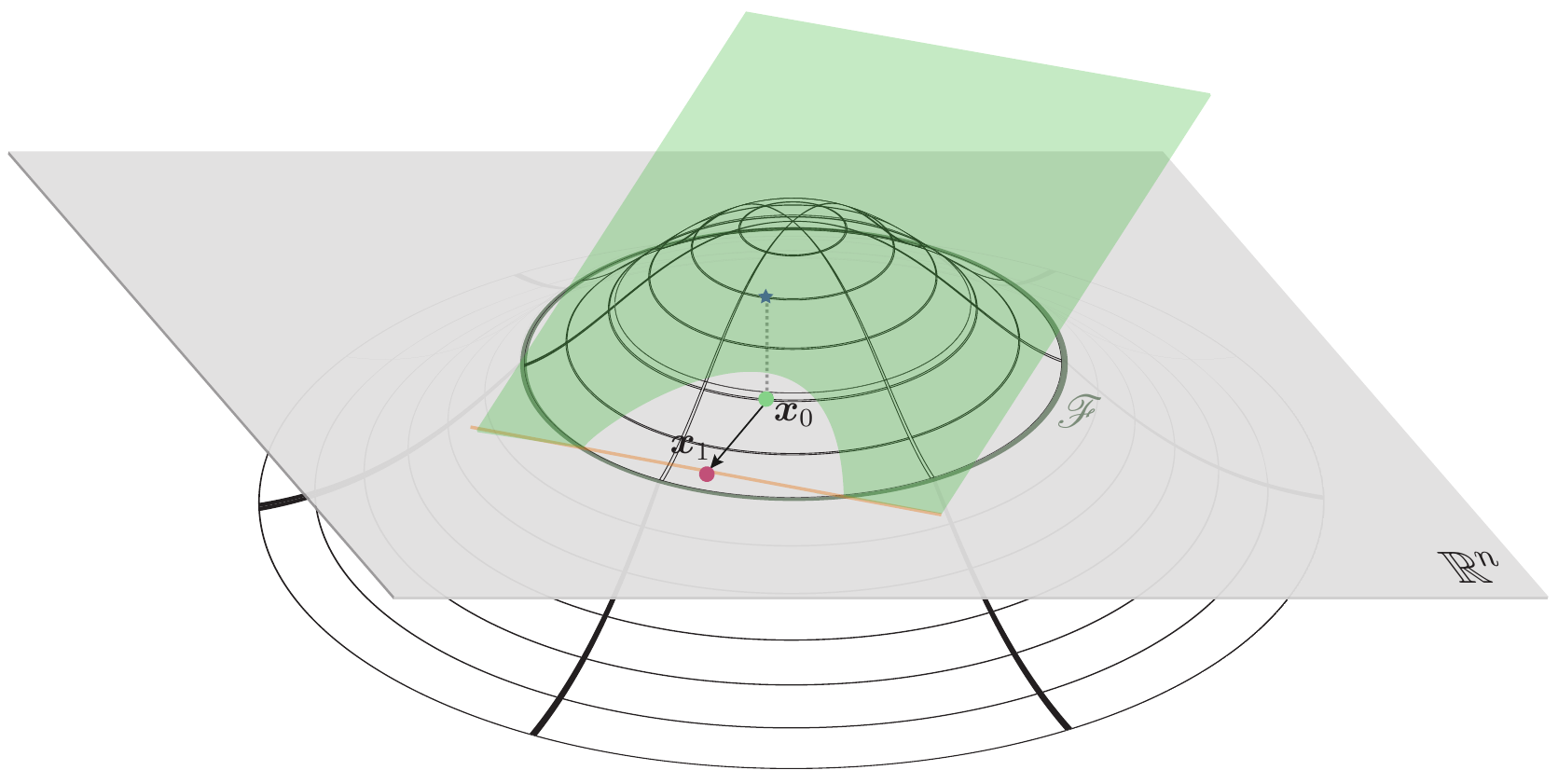}
\caption{\label{fig:binary_nonlin}Illustration of Algorithm \ref{alg:binary} for $n=2$. Assume $\x_0\in\mathbb{R}^n$. The green plane is the graph of $\x \mapsto f(\x_0)+\nabla f(\x_0)^T(\x-\x_0)$, which is tangent to the classifier function (wire-framed graph) $\x \mapsto f(\x)$. The orange line indicates where $f(\x_0)+\nabla f(\x_0)^T(\x-\x_0)=0$. $\x_1$ is obtained from $\x_0$ by projecting $\x_0$ on the orange hyperplane of $\mathbb{R}^n$.}
\end{figure}

In the case where the classifier $f$ is affine, it can easily be seen that the robustness of $f$ at point $\x_0$, $\Delta(\x_0; f)$\footnote{From now on, we refer to a classifier either by $f$ or its corresponding discrete mapping $\hat{k}$. Therefore, $\rho_{\text{adv}}(\hat{k})=\rho_{\text{adv}}(f)$ and $\Delta(\x;\hat{k})=\Delta(\x;f)$.}, is equal to the distance from $\x_0$ to the separating affine hyperplane $\mathscr{F} = \{ \x: \w^T\x+b= 0 \}$ (Figure \ref{fig:binary_lin}). The minimal perturbation to change the classifier's decision corresponds to the orthogonal projection of $\x_0$ onto $\mathscr{F}$. It is given by the closed-form formula:
\begin{align}
  \label{eq:update_linear}
\r_*(\x_0) & := \argmin \| \r \|_2 \\ & \text{ subject to } \text{ sign } (f(\x_0+\r)) \neq \text{ sign} (f(\x_0)) \nonumber \\ 
			  & = -\frac{f(\x_0)}{\|\w\|_2^2}\w.\nonumber			  			
\end{align}
Assuming now that $f$ is a general binary differentiable classifier, we adopt an iterative procedure to estimate the robustness $\Delta(\x_0; f)$. Specifically, at each iteration, $f$ is linearized around the current point $\x_i$ and the minimal perturbation of the linearized classifier is computed as
\begin{align}
\argmin_{\r_i} \| \r_i \|_2 \text{ subject to } f(\x_i) + \nabla f(\x_i)^T \r_i = 0.
\end{align}
The perturbation $\r_i$ at iteration $i$ of the algorithm is computed using the closed form solution in Eq. (\ref{eq:update_linear}), and the next iterate $\x_{i+1}$ is updated. The algorithm stops when $\x_{i+1}$ changes sign of the classifier. The DeepFool algorithm for binary classifiers is summarized in Algorithm \ref{alg:binary} and a geometric illustration of the method is shown in Figure \ref{fig:binary_nonlin}.


\begin{algorithm}
\caption{DeepFool for binary classifiers}
\begin{algorithmic}[1]
\State \textbf{input:} Image $\x$, classifier $f$.
\State \textbf{output:} Perturbation $\hat{\r}$.
\State Initialize $\x_0 \gets \x$, $i \gets 0$.
\While{$\text{sign} (f(\x_i)) = \text{sign} (f(\x_0))$}
\State $\r_i \gets -\frac{f(\x_i)}{\|\nabla f(x_i)\|_2^2}\nabla f(\x_i)$,
\State $\x_{i+1} \gets \x_i + \r_i$,
\State $i \gets i+1$.
\EndWhile
\State \textbf{return} $\hat{\r} = \sum_i \r_i$.
\end{algorithmic}
\label{alg:binary}
\end{algorithm}
In practice, the above algorithm can often converge to a point on the zero level set $\mathscr{F}$. In order to reach the other side of the classification boundary, the final perturbation vector $\hat{\r}$ is multiplied by a constant $1+\eta$, with $\eta \ll 1$. In our experiments, we have used $\eta=0.02$. 

\red{}

\section{DeepFool for multiclass classifiers}
We now extend the DeepFool method to the multiclass case. The most common used scheme for multiclass classifiers is one-vs-all. Hence, we also propose our method based on this classification scheme. In this scheme, the classifier has $c$ outputs where $c$ is the number of classes. Therefore, a classifier can be defined as $f:\mathbb{R}^n\rightarrow\mathbb{R}^c$ and the classification is done by the following mapping:
\begin{equation}
\label{eq:argmax_classifier}
\hat{k}(\x)=\argmax_k f_k(\x),
\end{equation}
where $f_k(\x)$ is the output of $f(\x)$ that corresponds to the $k^{\text{th}}$ class.
Similarly to the binary case, we first present the proposed approach for the linear case and then we generalize it to other classifiers.
\subsection{Affine multiclass classifier}
\begin{figure}
\center
\includegraphics[scale=0.7]{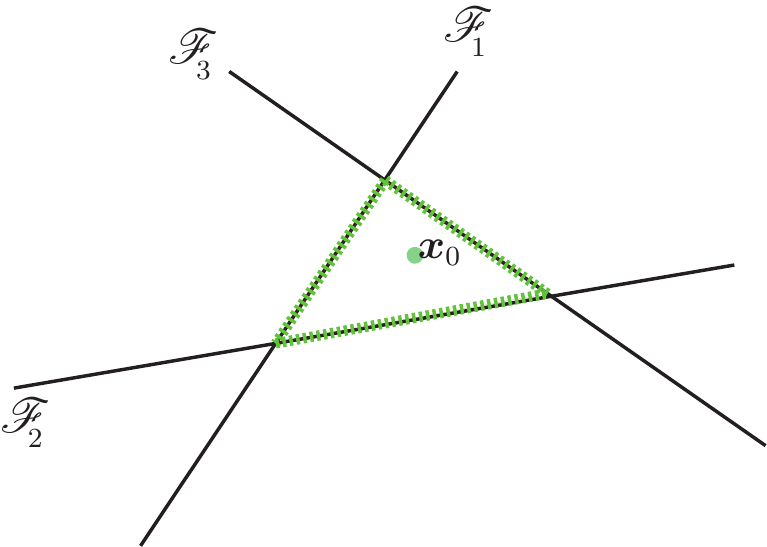}
\caption{For $\x_0$ belonging to class $4$, let $\mathscr{F}_k=\{\x:f_k(\x)-f_4(\x)=0\}$. These hyperplanes are depicted in solid lines and the boundary of $P$ is shown in green dotted line.}
\label{fig:multi_lin}
\end{figure}

Let $f(\x)$ be an affine classifier, i.e., $f(\x)=\mathbf{W}^\top\x+\boldsymbol{b}$ for a given $\mathbf{W}$ and $\boldsymbol{b}$. Since the mapping $\hat{k}$ is the outcome of a one-vs-all classification scheme, the minimal perturbation to fool the classifier can be rewritten as follows
\begin{equation}
\label{eq:prob_proj_compl_poly}
\begin{split}
&\argmin_{\r} \|\r\|_2\\
&\text{s.t. }\exists k: \w_k^\top(\x_0+\r)+b_k\geq\w_{\hat{k}(\x_0)}^\top(\x_0+\r)+b_{\hat{k}(\x_0)},
\end{split}
\end{equation}
where $\w_k$ is the $k^{\text{th}}$ column of $\mathbf{W}$. Geometrically, the above problem corresponds to the computation of the distance between $\x_0$ and the \textit{complement} of the convex polyhedron $P$,
\begin{align}
P = \bigcap_{k=1}^c \{ \x: f_{\hat{k} (\x_0)} (\x) \geq f_{k} (\x) \},
\label{eq:polyhedron_P}
\end{align}
where $\x_0$ is located inside $P$.
We denote this distance by $\text{\textbf{dist}}(\x_0,P^c)$.
The polyhedron $P$ defines the region of the space where $f$ outputs the label $\hat{k} (\x_0)$. This setting is depicted in Figure \ref{fig:multi_lin}. The solution to the problem in Eq. (\ref{eq:prob_proj_compl_poly}) can be computed in closed form as follows. Define $\hat{l}(\x_0)$ to be the closest hyperplane of the boundary of $P$ (e.g. $\hat{l}(\x_0)=3$ in Figure \ref{fig:multi_lin}). Formally, $\hat{l} (\x_0)$ can be computed as follows
\begin{equation}
\hat{l}(\x_0)=\argmin_{k\neq{\hat{k}(\x_0)}}\frac{\left|f_k(\x_0)-f_{\hat{k}(\x_0)}(\x_0)\right|}{\|\w_k-\w_{\hat{k}(\x_0)}\|_2}.
\end{equation}
The minimum perturbation $\r_*(\x_0)$ is the vector that projects $\x_0$ on the hyperplane indexed by $\hat{l}(\x_0)$, i.e.,
\begin{equation}
\r_*(\x_0)=\frac{\left|f_{\hat{l}(\x_0)}(\x_0)-f_{\hat{k}(\x_0)}(\x_0)\right|}{\|\w_{\hat{l}(\x_0)}-\w_{\hat{k}(\x_0)}\|_2^2}(\w_{\hat{l}(\x_0)}-\w_{\hat{k}(\x_0)}).
\end{equation}
In other words, we find the closest projection of $\x_0$ on faces of $P$.
\subsection{General classifier}
We now extend the DeepFool algorithm to the general case of multiclass differentiable classifiers. 
For general non-linear classifiers, the set $P$ in Eq. (\ref{eq:polyhedron_P}) that describes the region of the space where the classifier outputs label $\hat{k}(\x_0)$ is no longer a polyhedron. Following the explained iterative linearization procedure in the binary case, we approximate the set $P$ at iteration $i$ by a polyhedron $\tilde{P}_i$
\begin{align}
\tilde{P}_i=\bigcap_{k=1}^c\Big\{& \x:  f_k(\x_i)-f_{\hat{k}(\x_0)}(\x_i)\\
& + \nabla f_k(\x_i)^\top\x-\nabla f_{\hat{k}(\x_0)}(\x_i)^\top\x\leq 0\Big\}. \nonumber
\end{align}
%
%
We then approximate, at iteration $i$, the distance between $\x_i$ and the complement of $P$, $\text{\textbf{dist}}(\x_i,P^c)$, by $\text{\textbf{dist}}(\x_i,\tilde{P}_i^c)$. Specifically, at each iteration of the algorithm, the perturbation vector that reaches the boundary of the polyhedron $\tilde{P}_i$ is computed, and the current estimate updated.
 The method is given in Algorithm \ref{alg:multiclass}. It should be noted that the proposed algorithm operates in a greedy way and is not guaranteed to converge to the optimal perturbation in (\ref{eq:opt_robustness}). However, we have observed in practice that our algorithm yields very small perturbations which are believed to be good approximations of the minimal perturbation. 
 
\begin{figure}
\center
\includegraphics[scale=0.7]{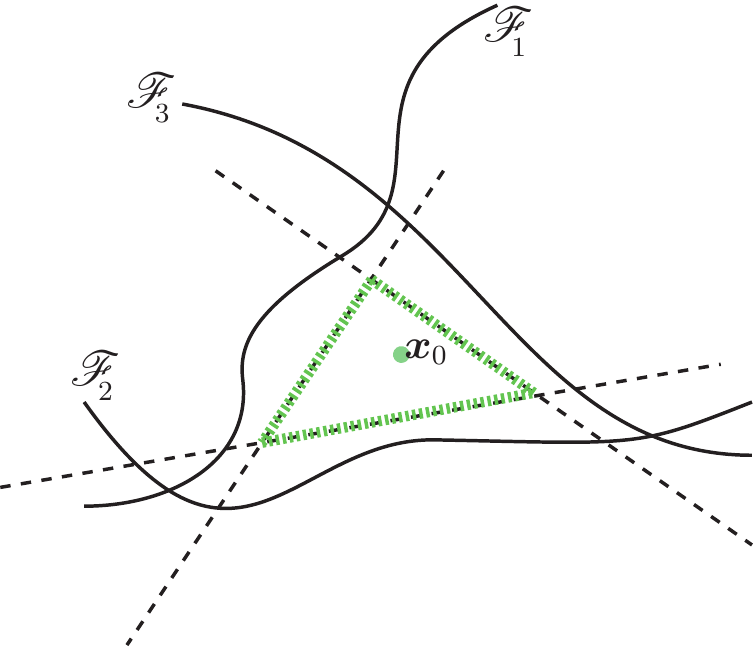}
\caption{For $\x_0$ belonging to class $4$, let $\mathscr{F}_k=\{\x:f_k(\x)-f_4(\x)=0\}$. The linearized zero level sets are shown in dashed lines and the boundary of the polyhedron $\tilde{P}_0$ in green.}
\label{fig:multi_general}
\end{figure}
\begin{algorithm}
\caption{DeepFool: multi-class case}
\begin{algorithmic}[1]
\State \textbf{input:} Image $\x$, classifier $f$.
\State \textbf{output:} Perturbation $\hat{\r}$.
\\
\State Initialize $\x_0\gets \x$, $i \gets 0$.
\While{$\hat{k}(\x_i)=\hat{k}(\x_0)$}
\For{$k\neq \hat{k}(\x_0)$}
	\State $\w'_k\gets \nabla f_k(\x_i)-\nabla f_{\hat{k}(\x_0)}(\x_i)$
	\State $f'_k\gets f_k(\x_i)-f_{\hat{k}(\x_0)}(\x_i)$
\EndFor
\State $\hat{l}\gets\argmin_{k\neq{\hat{k}(\x_0)}}\frac{\left|f'_k\right|}{\|\w'_k\|_2}$
\State $\r_i\gets \frac{\left|f'_{\hat{l}}\right|}{\|\w'_{\hat{l}}\|_2^2}\w'_{\hat{l}}$
\State $\x_{i+1}\gets\x_i+\r_i$
\State $i\gets i+1$
\EndWhile
\State \textbf{return} $\hat{\r} = \sum_i \r_i$
\end{algorithmic}
\label{alg:multiclass}
\end{algorithm}

It should be noted that the optimization strategy of DeepFool is strongly tied to existing optimization techniques. In the binary case, it can be seen as Newton's iterative algorithm for finding roots of a nonlinear system of equations in the underdetermined case \cite{ruszczynski2006}. This algorithm is known as the normal flow method. The convergence analysis of this optimization technique can be found for example in \cite{walker90}. Our algorithm in the binary case can alternatively be seen as a gradient descent algorithm with an adaptive step size that is automatically chosen at each iteration. The linearization in Algorithm \ref{alg:multiclass} is also similar to a sequential convex programming where the constraints are linearized at each step.

\subsection{Extension to $\ell_p$ norm}
\label{sec:p_norm}
In this paper, we have measured the perturbations using the $\ell_2$ norm. Our framework is however not limited to this choice, and the proposed algorithm can simply be adapted to find minimal adversarial perturbations for any $\ell_p$ norm ($p \in [1,\infty)$). To do so, the update steps in line 10 and 11 in Algorithm \ref{alg:multiclass} must be respectively substituted by the following updates
\begin{align}
\hat{l} &\gets\argmin_{k\neq{\hat{k}(\x_0)}}\frac{\left|f'_k\right|}{\|\w'_k\|_q}, \\
\r_i &\gets \frac{|f'_{\hat{l}}|}{\|\w'_{\hat{l}}\|_q^q}|\w'_{\hat{l}}|^{q-1}\odot\text{sign}(\w'_{\hat{l}}),
\end{align}
where $\odot$ is the pointwise product and $q=\frac{p}{p-1}$.\footnote{To see this, one can apply Holder's inequality to obtain a lower bound on the $\ell_p$ norm of the perturbation.} In particular, when $p = \infty$ (i.e., the supremum norm $\ell_{\infty}$), these update steps become
\begin{align}
\hat{l} &\gets\argmin_{k\neq{\hat{k}(\x_0)}}\frac{\left|f'_k\right|}{\|\w'_k\|_1}, \\
\r_i &\gets \frac{|f'_{\hat{l}}|}{\|\w'_{\hat{l}}\|_1}\text{sign}(\w'_{\hat{l}}).
\end{align}
\section{Experimental results}
\label{sec:eval}
\subsection{Setup}
We now test our DeepFool algorithm on deep convolutional neural networks architectures applied to MNIST, CIFAR-10, and ImageNet image classification datasets. We consider the following deep neural network architectures:
\begin{itemize}
\item \textbf{MNIST:} A two-layer fully connected network, and a two-layer LeNet convoluational neural network architecture \cite{lecun99}. Both networks are trained with SGD with momentum using the MatConvNet \cite{vedaldi2015} package.
\item \textbf{CIFAR-10:} We trained a three-layer LeNet architecture, as well as a Network In Network (NIN) architecture \cite{lin2013}.
\item \textbf{ILSVRC 2012:} We used CaffeNet \cite{jia2014} and GoogLeNet \cite{szegedy2015} pre-trained models.
\end{itemize}

In order to evaluate the robustness to adversarial perturbations of a classifier $f$, we compute the average robustness $\hat{\rho}_{\text{adv}}(f)$, defined by 
\begin{equation}
\hat{\rho}_{\text{adv}}(f)=\frac{1}{|\mathscr{D}|} \sum_{\x\in\mathscr{D}}\frac{\| \hat{\r} (\x) \|_2 }{\|\x\|_2},
\label{eq:rho_adv}
\end{equation}
where $\hat{\r}(\x)$ is the estimated minimal perturbation obtained using DeepFool, and $\mathscr{D}$ denotes the test set\footnote{For ILSVRC2012, we used the validation data.}.

We compare the proposed DeepFool approach to state-of-the-art techniques to compute adversarial perturbations in \cite{szegedy2013} and \cite{goodfellow2014}. The method in \cite{szegedy2013} solves a series of penalized optimization problems to find the minimal perturbation, whereas \cite{goodfellow2014} estimates the minimal perturbation by taking the sign of the gradient
\begin{align*}
\hat{\r} (\x) = \epsilon\,\text{sign} \left( \nabla_{\x} J(\boldsymbol{\theta}, \x, y) \right),
\end{align*} 
with $J$ the cost used to train the neural network, $\boldsymbol{\theta}$ is the model parameters, and $y$ is the label of $\x$. The method is called \textit{fast gradient sign method}. In practice, in the absence of general rules to choose the parameter $\epsilon$, we chose the smallest $\epsilon$ such that $90\%$ of the data are misclassified after perturbation.\footnote{Using this method, we observed empirically that one cannot reach $100\%$ misclassification rate on some datasets. In fact, even by increasing $\epsilon$ to be very large, this method can fail in misclassifying all samples.}



\subsection{Results}
We report in Table \ref{tab:mean_pert} the accuracy and average robustness $\hat{\rho}_{\text{adv}}$ of each classifier computed using different methods. We also show the running time required for each method to compute \textit{one} adversarial sample.
%
\begin{table*}[]
\centering
\renewcommand{\arraystretch}{1.5}
\begin{tabular}{|l|l|l|l|l|l|l|l|l|l}
\cline{1-8}
\textbf{Classifier}    & \textbf{Test error} & $\hat{\rho}_\text{adv}$ [DeepFool] &time& $\hat{\rho}_\text{adv}$ \cite{goodfellow2014} &time& $\hat{\rho}_\text{adv}$ \cite{szegedy2013}&time\\ \cline{1-8}
LeNet (MNIST) & 1\%      & $2.0\times 10^{-1}$  &110 ms  & 1.0   & 20 ms& $2.5\times 10^{-1}$  & $>$ 4 s\\ \cline{1-8}
FC500-150-10 (MNIST)& 1.7\%      & $1.1\times 10^{-1}$      & 50 ms & $3.9\times 10^{-1}$       & 10 ms &$1.2\times 10^{-1}$ &$>$ 2 s \\ \cline{1-8}
NIN (CIFAR-10)   & 11.5\%      & $2.3\times 10^{-2}$    &1100 ms& $1.2\times 10^{-1}$     &180 ms& $2.4\times 10^{-2}$     &$>$50 s\\ \cline{1-8}
LeNet (CIFAR-10)  & 22.6\%      & $3.0\times 10^{-2}$    &220 ms & $1.3\times 10^{-1}$   &50 ms& $3.9\times 10^{-2}$   & $>$7 s \\ \cline{1-8}
CaffeNet (ILSVRC2012)&   42.6\%    & $2.7\times 10^{-3}$      &510 ms*& $3.5\times 10^{-2}$  &50 ms*&-&-\\ \cline{1-8}
GoogLeNet (ILSVRC2012)  &    31.3\%   & $1.9\times 10^{-3}$    &800 ms*&$4.7\times 10^{-2}$     & 80 ms*&-&-\\ \cline{1-8}
\end{tabular}
\vspace{10pt}
\caption{The adversarial robustness of different classifiers on different datasets. The time required to compute one sample for each method is given in the time columns. The times are computed on a Mid-2015 MacBook Pro without CUDA support. The asterisk marks determines the values computed using a GTX 750 Ti GPU.}
\label{tab:mean_pert}
\end{table*}
It can be seen that DeepFool estimates smaller perturbations (hence closer to minimal perturbation defined in (\ref{eq:opt_robustness})) than the ones computed using the competitive approaches. For example, the average perturbation obtained using DeepFool is $5$ times lower than the one estimated with \cite{goodfellow2014}. On the ILSVRC2012 challenge dataset, the average perturbation is one order of magnitude smaller compared to the fast gradient method. It should be noted moreover that the proposed approach also yields slightly smaller perturbation vectors than the method in \cite{szegedy2013}. The proposed approach is hence more accurate in detecting directions that can potentially fool neural networks. As a result, DeepFool can be used as a valuable tool to accurately assess the robustness of classifiers. 
On the complexity aspect, the proposed approach is substantially faster than the standard method proposed in \cite{szegedy2013}.
In fact, while the approach \cite{szegedy2013} involves a costly minimization of a series of objective functions, we observed empirically that DeepFool converges in a few iterations (i.e., less than $3$) to a perturbation vector that fools the classifier. Hence, the proposed approach reaches a more accurate perturbation vector compared to state-of-the-art methods, while being computationally efficient. This makes it readily suitable to be used as a baseline method to estimate the robustness of very deep neural networks on large-scale datasets. In that context, we provide the first quantitative evaluation of the robustness of state-of-the-art classifiers on the large-scale ImageNet dataset. It can be seen that despite their very good test accuracy, these methods are extremely unstable to adversarial perturbations: a perturbation that is $1000$ smaller in magnitude than the original image is sufficient to fool state-of-the-art deep neural networks.



We illustrate in Figure \ref{fig:perturbed_smpl} perturbed images generated by the fast gradient sign and DeepFool. It can be observed that the proposed method generates adversarial perturbations which are hardly perceptible, while the fast gradient sign method outputs a perturbation image with higher norm.

It should be noted that, when perturbations are measured using the $\ell_{\infty}$ norm, the above conclusions remain unchanged: DeepFool yields adversarial perturbations that are smaller (hence closer to the optimum) compared to other methods for computing adversarial examples. Table \ref{tab:l_infty} reports the $\ell_{\infty}$ robustness to adversarial perturbations measured by  $\hat{\rho}_{\text{adv}}^\infty(f)=\frac{1}{|\mathscr{D}|} \sum_{\x\in\mathscr{D}}\frac{\| \hat{\r} (\x) \|_\infty }{\|\x\|_\infty}$, where $\hat{\r} (\x)$ is computed respectively using DeepFool (with $p=\infty$, see Section \ref{sec:p_norm}), and the Fast gradient sign method for MNIST and CIFAR-10 tasks.

\begin{table}[t]
\fontsize{8pt}{15pt}
\selectfont
\begin{center}
\begin{tabular}{|l|c|c|}
\hline
 \textbf{Classifier}         & DeepFool &Fast gradient sign\\ \hline
LeNet (MNIST) &0.10    & 0.26      \\ \hline
FC500-150-10 (MNIST)   &  0.04  & 0.11         \\ \hline
NIN (CIFAR-10)&0.008& 0.024\\ \hline
LeNet (CIFAR-10)&0.015&0.028\\ \hline
\end{tabular}
\end{center}
\caption{Values of $\hat{\rho}_{\text{adv}}^\infty$ for four different networks based on DeepFool (smallest $l_\infty$ perturbation) and fast gradient sign method with 90\% of misclassification.}
\label{tab:l_infty}
\end{table}

\begin{figure}[h]
\centering
\includegraphics[scale=0.3]{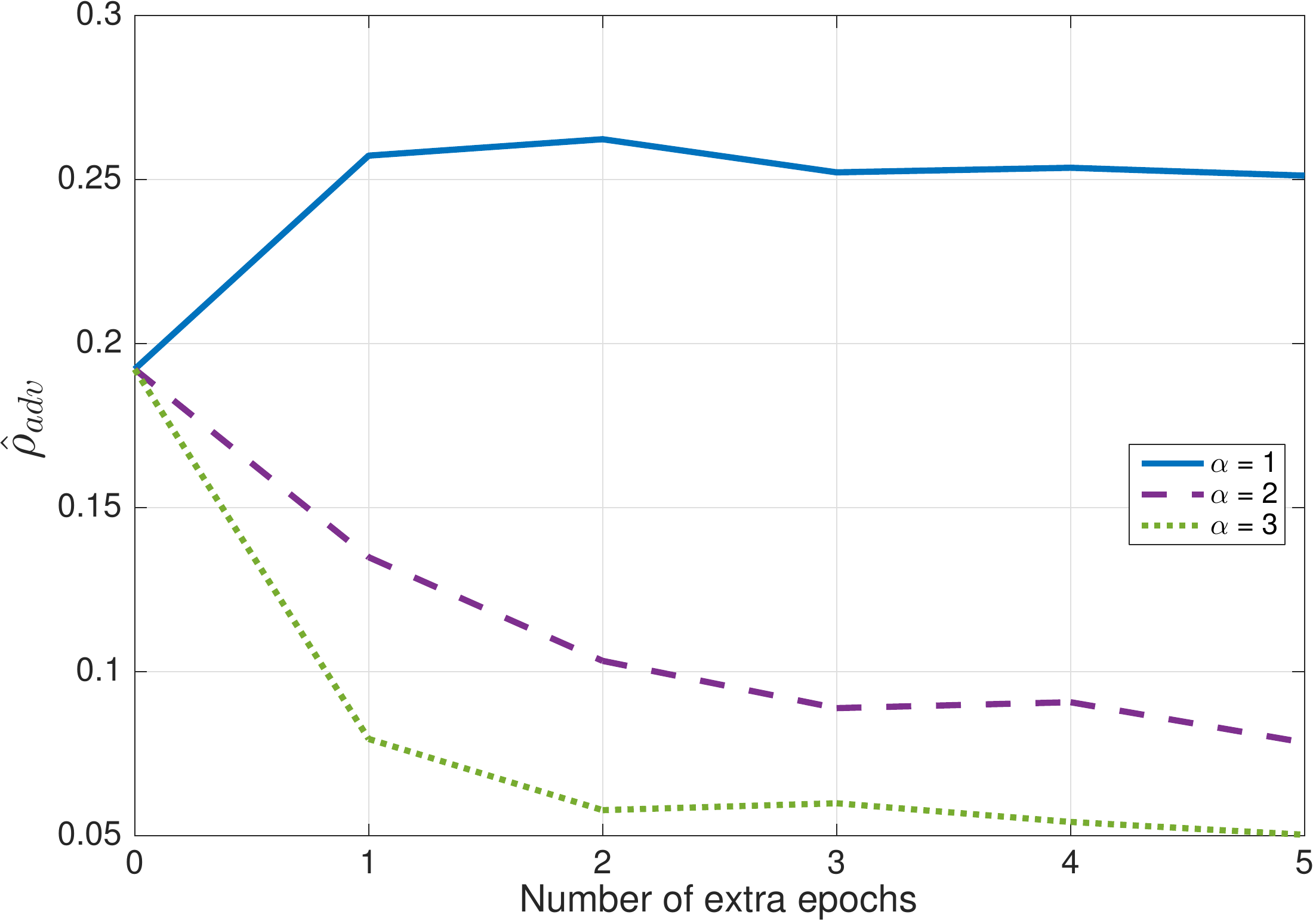}
\caption{Fine-tuning based on magnified DeepFool's adversarial perturbations.}
\label{fig:mult_pert}
\end{figure}
\begin{figure}[h]
\centering
\includegraphics[scale=0.4]{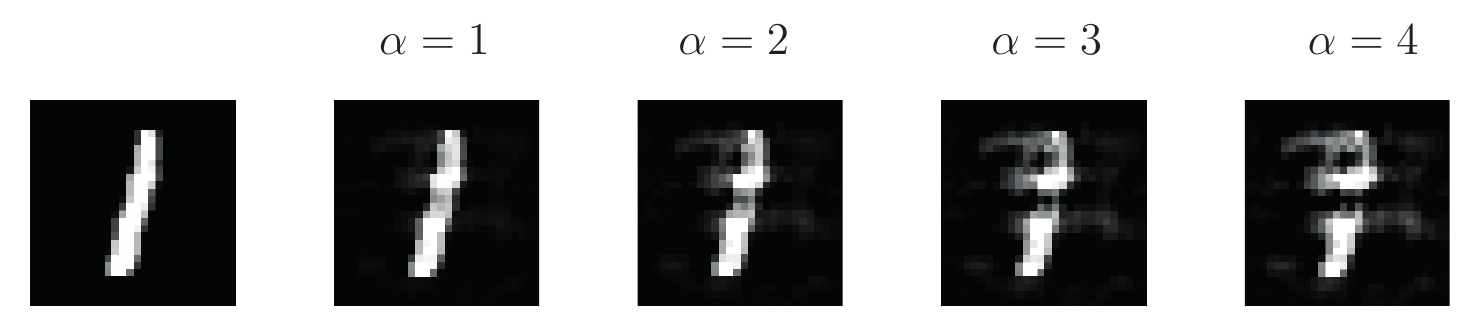}
\caption{From ``1" to ``7"	: original image classified as ``1" and the DeepFool perturbed images classified as ``7" using different values of $\alpha$.}
\label{fig:perturbed_smpl2}
\end{figure}

\textbf{Fine-tuning using adversarial examples}
\begin{figure*}[ht]
\centering
\begin{subfigure}[t]{200pt}
\includegraphics[scale=0.3]{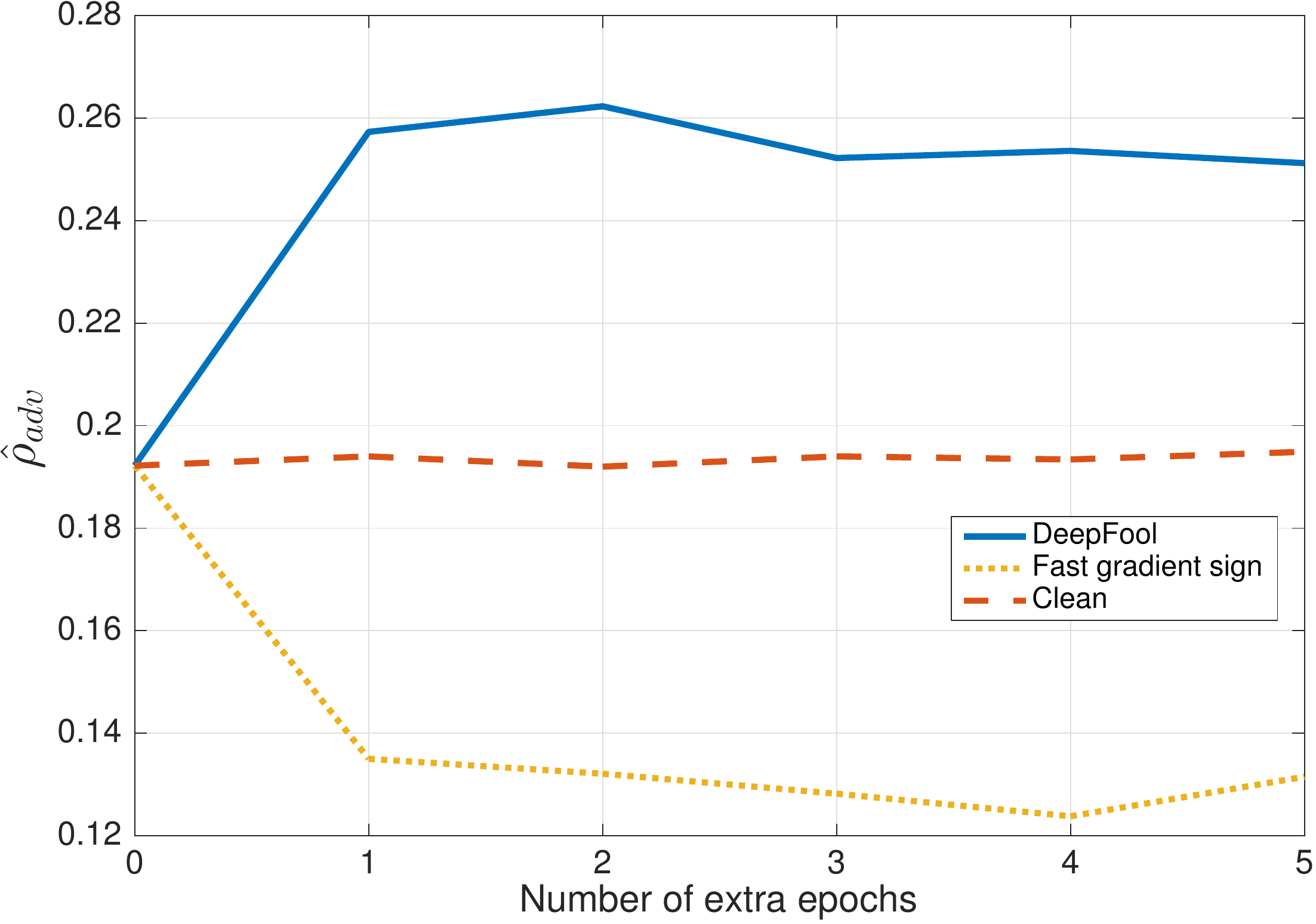}
\caption{Effect of fine-tuning on adversarial examples computed by two different methods for LeNet on MNIST.}
\label{tab:mnist_back_feed1}
\end{subfigure}
~
\begin{subfigure}[t]{200pt}
\includegraphics[scale=0.3]{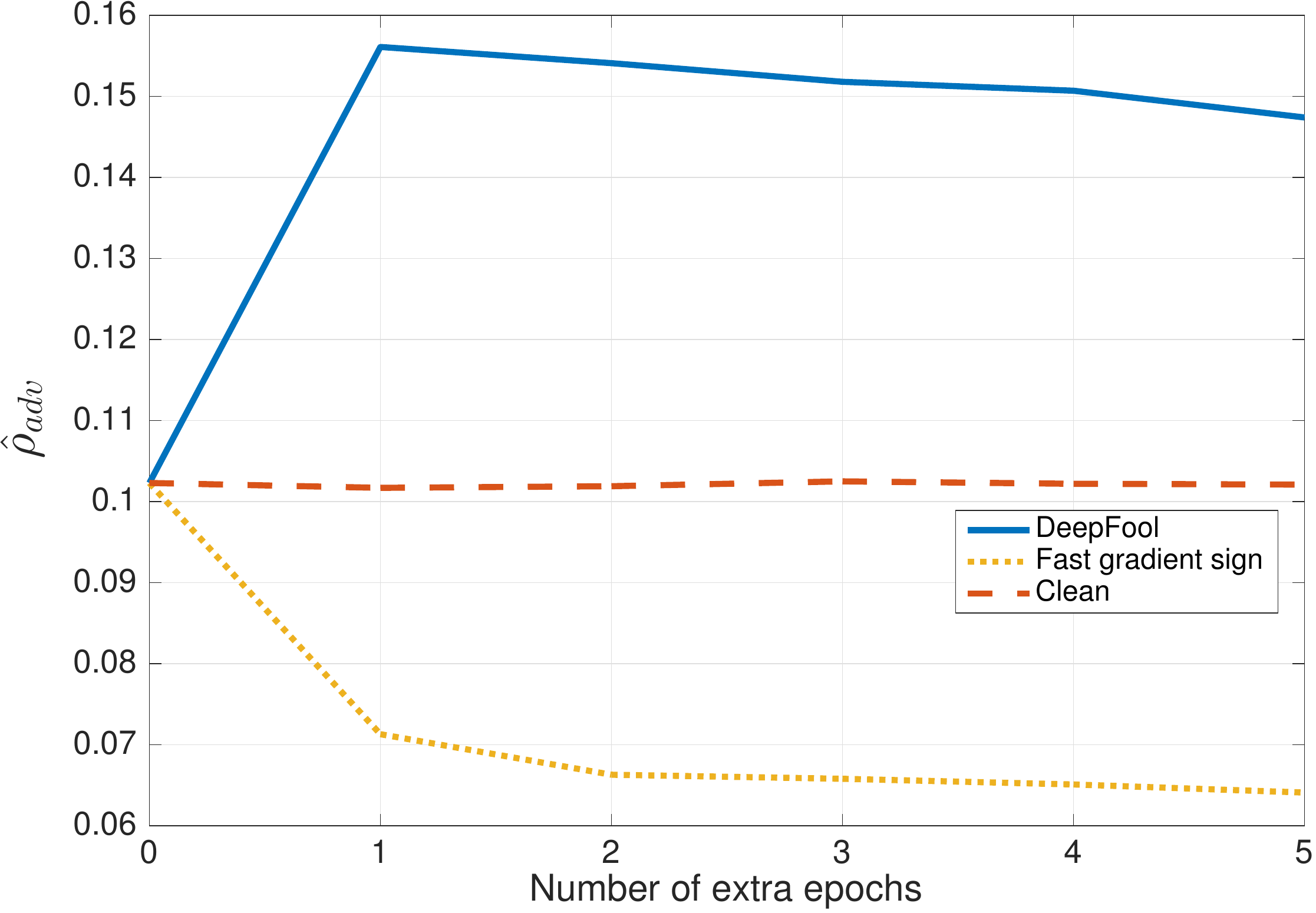}
\caption{Effect of fine-tuning on adversarial examples computed by two different methods for a fully-connected network on MNIST.}
\label{tab:mnist_back_feed2}
\end{subfigure}

\begin{subfigure}[t]{200pt}
\includegraphics[scale=0.3]{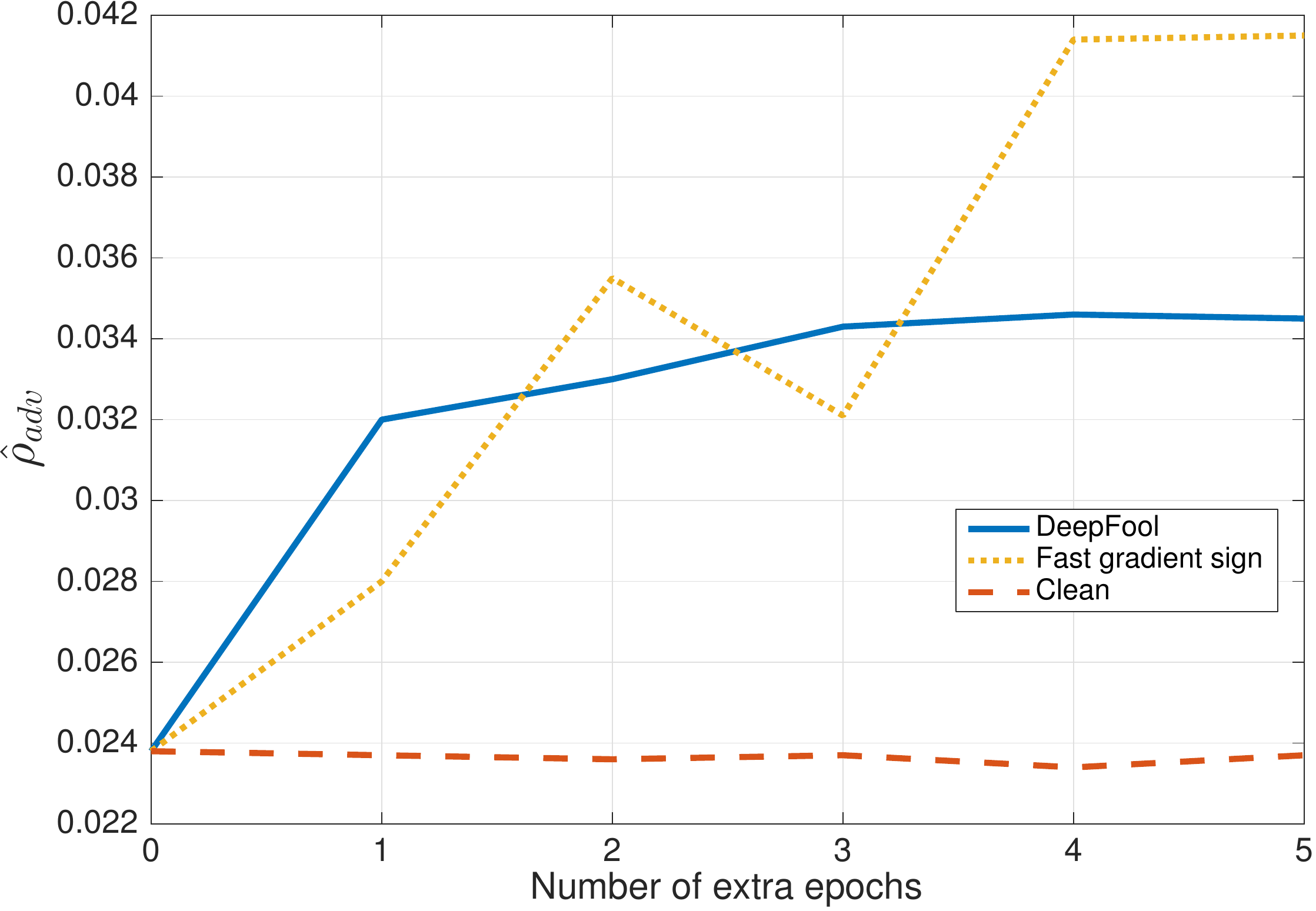}
\caption{Effect of fine-tuning on adversarial examples computed by two different methods for NIN on CIFAR-10.}
\label{tab:cifar_back_feed1}
\end{subfigure}
~
\begin{subfigure}[t]{200pt}
\includegraphics[scale=0.3]{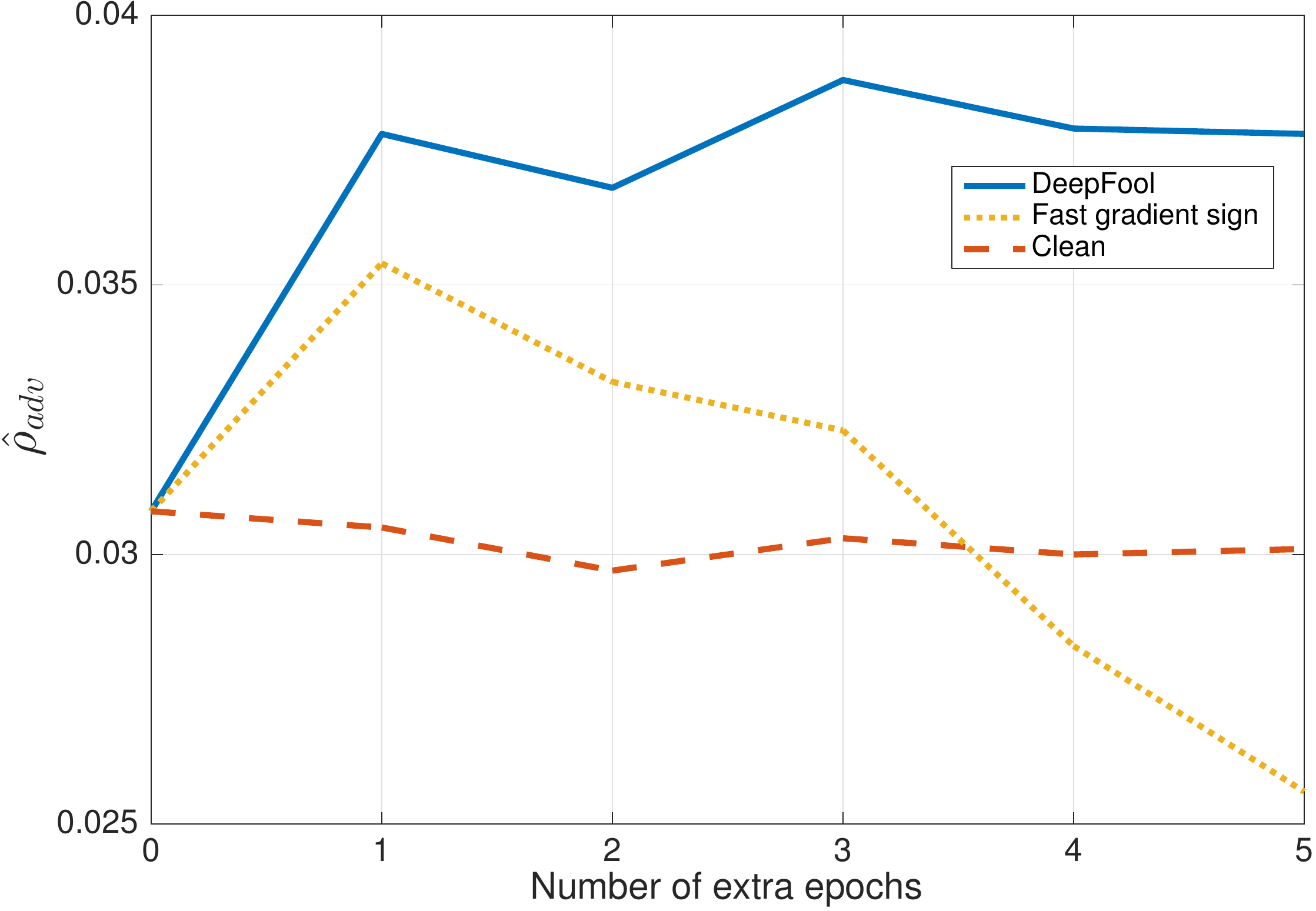}
\caption{Effect of fine-tuning on adversarial examples computed by two different methods for LeNet on CIFAR-10.}
\label{tab:cifar_back_feed2}
\end{subfigure}
\caption{\label{fig:feedbacked}}
\end{figure*}
In this section, we fine-tune the networks of Table \ref{tab:mean_pert} on adversarial examples to build more robust classifiers for the MNIST and CIFAR-10 tasks. Specifically, for each network, we performed two experiments: (i) Fine-tuning the network on DeepFool's adversarial examples, (ii) Fine-tuning the network on the fast gradient sign adversarial examples.
We fine-tune the networks by performing 5 additional epochs, with a $50\%$ decreased learning rate only on the perturbed training set. For each experiment, the same training data was used through all $5$ extra epochs. For the sake of completeness, we also performed $5$ extra epochs on the original data. The evolution of $\hat{\rho}_{\text{adv}}$ for the different fine-tuning strategies is shown in Figures \ref{tab:mnist_back_feed1} to \ref{tab:cifar_back_feed2}, \textit{where the robustness $\hat{\rho}_{\text{adv}}$ is estimated using DeepFool}, since this is the most accurate method, as shown in Table \ref{tab:mean_pert}. Observe that fine-tuning with DeepFool adversarial examples significantly increases the robustness of the networks to adversarial perturbations even after one extra epoch. For example, the robustness of the networks on MNIST is improved by 50\% and NIN's robustness is increased by about 40\%. On the other hand, quite surprisingly, the method in \cite{goodfellow2014} can lead to \textit{a decreased} robustness to adversarial perturbations of the network. We hypothesize that this behavior is due to the fact that perturbations estimated using the fast gradient sign method are much larger than minimal adversarial perturbations. Fine-tuning the network with overly perturbed images decreases the robustness of the networks to adversarial perturbations. 
To verify this hypothesis, we compare in Figure \ref{fig:mult_pert} the adversarial robustness of a network that is fine-tuned with the adversarial examples obtained using DeepFool, where norms of perturbations have been deliberately multiplied by $\alpha=1,2,3$. Interestingly, we see that by magnifying the norms of the adversarial perturbations, the robustness of the fine-tuned network is \textit{decreased}. This might explain why overly perturbed images decrease the robustness of MNIST networks: these perturbations can really change the class of the digits, hence fine-tuning based on these examples can lead to a drop of the robustness (for an illustration, see Figure \ref{fig:perturbed_smpl2}). This lends credence to our hypothesis, and further shows the importance of designing accurate methods to compute minimal perturbations.

Table \ref{tab:acc_} lists the accuracies of the fine-tuned networks. It can be seen that fine-tuning with DeepFool can improve the accuracy of the networks. Conversely, fine-tuning with the approach in \cite{goodfellow2014} has led to a \textit{decrease} of the test accuracy in all our experiments. This confirms the explanation that the fast gradient sign method outputs \textit{overly perturbed} images that lead to images that are unlikely to occur in the test data. Hence, it \textit{decreases} the performance of the method as it acts as a regularizer that does not represent the distribution of the original data. This effect is analogous to geometric data augmentation schemes, where \textit{large} transformations of the original samples have a counter-productive effect on generalization.\footnote{While the authors of \cite{goodfellow2014} reported an \textit{increased} generalization performance on the MNIST task (from $0.94\%$ to $0.84\%$) using adversarial regularization, it should be noted that the their experimental setup is significantly different as \cite{goodfellow2014} trained the network based on a modified cost function, while we performed straightforward fine-tuning.}



\begin{figure}[]
\center
\includegraphics[scale=0.3]{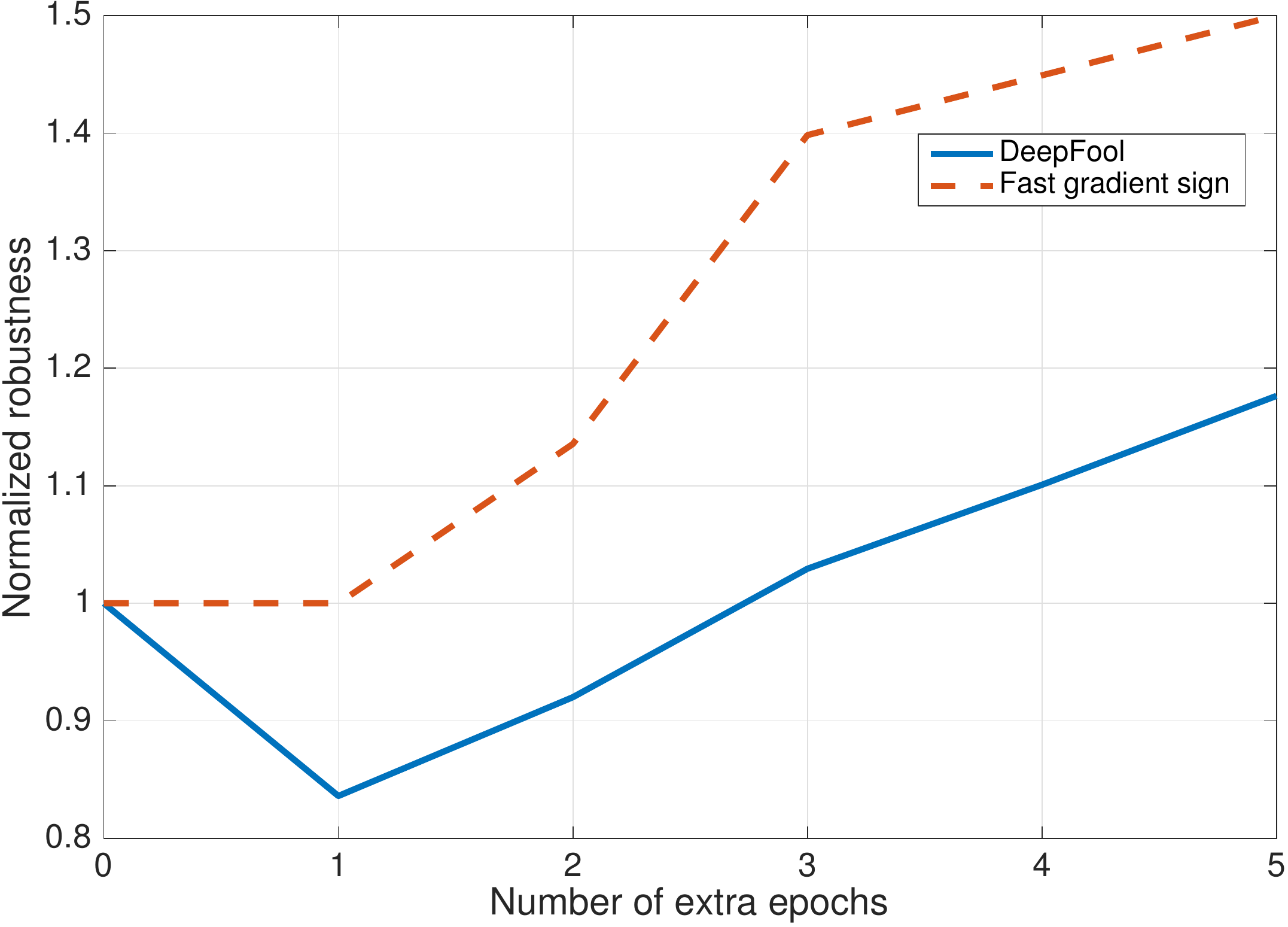}
\caption{How the adversarial robustness is judged by different methods. The values are normalized by the corresponding $\hat{\rho}_{\text{adv}}$s of the original network.}
\label{fig:nin_back_feed_signofgradient}
\end{figure}

To emphasize the importance of a correct estimation of the minimal perturbation, we now show that using approximate methods can lead to wrong conclusions regarding the adversarial robustness of networks. We fine-tune the NIN classifier on the fast gradient sign adversarial examples. We follow the procedure described earlier but this time, we decreased the learning rate by 90\%. We have evaluated the adversarial robustness of this network at different extra epochs using DeepFool and the \textit{fast gradient sign method}. As one can see in Figure \ref{fig:nin_back_feed_signofgradient}, the red plot exaggerates the effect of training on the adversarial examples. Moreover, it is not sensitive enough to demonstrate the loss of robustness at the first extra epoch. These observations confirm that using an \textit{accurate} tool to measure the robustness of classifiers is crucial to derive conclusions about the robustness of networks. 

\begin{table}[]
\fontsize{8pt}{15pt}
\selectfont
\begin{center}
\begin{tabular}{|l|c|c|c|}
\hline
 \textbf{Classifier}         & DeepFool &Fast gradient sign & Clean\\ \hline
LeNet (MNIST) &0.8\%    & 4.4\%    & 1\%      \\ \hline
FC500-150-10 (MNIST)   & 1.5\%   & 4.9\% &1.7\%         \\ \hline
NIN (CIFAR-10)&11.2\%&21.2\%&11.5\%\\ \hline
LeNet (CIFAR-10)&20.0\%&28.6\%&22.6\%\\ \hline
\end{tabular}
\end{center}
\caption{The test error of networks after the fine-tuning on adversarial examples (after five epochs). Each columns correspond to a different type of augmented perturbation. }
\label{tab:acc_}
\end{table}


%

\section{Conclusion}
In this work, we proposed an algorithm, DeepFool, to compute adversarial examples that fool state-of-the-art classifiers. It is based on an iterative linearization of the classifier to generate minimal perturbations that are sufficient to change classification labels. We provided extensive experimental evidence on three datasets and eight classifiers, showing the superiority of the proposed method over state-of-the-art methods to compute adversarial perturbations, as well as the efficiency of the proposed approach. Due to its accurate estimation of the adversarial perturbations, the proposed DeepFool algorithm provides an efficient and accurate way to evaluate the robustness of classifiers and to enhance their performance by proper fine-tuning.
The proposed approach can therefore be used as a reliable tool to accurately estimate the minimal perturbation vectors, and build more robust classifiers.
\subsection*{Acknowledgements}
This work has been partly supported by the Hasler Foundation, Switzerland, in the framework of the CORA project.
{\small
\bibliographystyle{ieee}
\bibliography{egbib}

\begin{thebibliography}{10}\itemsep=-1pt

\bibitem{bio1}
D.~Chicco, P.~Sadowski, and P.~Baldi.
\newblock Deep autoencoder neural networks for gene ontology annotation
  predictions.
\newblock In {\em ACM Conference on Bioinformatics, Computational Biology, and
  Health Informatics}, pages 533--540, 2014.

\bibitem{fawzi2015a}
A.~Fawzi, O.~Fawzi, and P.~Frossard.
\newblock Analysis of classifiers' robustness to adversarial perturbations.
\newblock {\em CoRR}, abs/1502.02590, 2015.

\bibitem{fawzi2015b}
A.~Fawzi and P.~Frossard.
\newblock Manitest: Are classifiers really invariant?
\newblock In {\em British Machine Vision Conference (BMVC)}, pages
  106.1--106.13, 2015.

\bibitem{goodfellow2014}
I.~J. Goodfellow, J.~Shlens, and C.~Szegedy.
\newblock Explaining and harnessing adversarial examples.
\newblock In {\em International Conference on Learning Representations}, 2015.

\bibitem{gu2015}
S.~Gu and L.~Rigazio.
\newblock Towards deep neural network architectures robust to adversarial
  examples.
\newblock {\em CoRR}, abs/1412.5068, 2014.

\bibitem{sp2}
G.~E. Hinton, L.~Deng, D.~Yu, G.~E. Dahl, A.~Mohamed, N.~Jaitly, A.~Senior,
  V.~Vanhoucke, P.~Nguyen, T.~N. Sainath, and B.~Kingsbury.
\newblock Deep neural networks for acoustic modeling in speech recognition: The
  shared views of four research groups.
\newblock {\em IEEE Signal Process. Mag.}, 29(6):82--97, 2012.

\bibitem{jia2014}
Y.~Jia, E.~Shelhamer, J.~Donahue, S.~Karayev, J.~Long, R.~Girshick,
  S.~Guadarrama, and T.~Darrell.
\newblock Caffe: Convolutional architecture for fast feature embedding.
\newblock In {\em ACM International Conference on Multimedia (MM)}, pages
  675--678. ACM, 2014.

\bibitem{cv2}
A.~Krizhevsky, I.~Sutskever, and G.~E. Hinton.
\newblock Imagenet classification with deep convolutional neural networks.
\newblock In {\em Advances in neural information processing systems (NIPS)},
  pages 1097--1105, 2012.

\bibitem{lecun99}
Y.~LeCun, P.~Haffner, L.~Bottou, and Y.~Bengio.
\newblock Object recognition with gradient-based learning.
\newblock In {\em Shape, contour and grouping in computer vision}, pages
  319--345. 1999.

\bibitem{cv1}
Y.~LeCun, K.~Kavukcuoglu, C.~Farabet, et~al.
\newblock Convolutional networks and applications in vision.
\newblock In {\em IEEE International Symposium on Circuits and Systems
  (ISCAS)}, pages 253--256, 2010.

\bibitem{lin2013}
M.~Lin, Q.~Chen, and S.~Yan.
\newblock Network in network.
\newblock 2014.

\bibitem{sp1}
T.~Mikolov, A.~Deoras, D.~Povey, L.~Burget, and J.~{\v{C}}ernock{\`y}.
\newblock Strategies for training large scale neural network language models.
\newblock In {\em IEEE Workshop on Automatic Speech Recognition and
  Understanding (ASRU)}, pages 196--201, 2011.

\bibitem{nguyen2015}
A.~Nguyen, J.~Yosinski, and J.~Clune.
\newblock Deep neural networks are easily fooled: High confidence predictions
  for unrecognizable images.
\newblock In {\em IEEE Conference on Computer Vision and Pattern Recognition
  (CVPR)}, pages 427--436, 2015.

\bibitem{pepik2015}
B.~Pepik, R.~Benenson, T.~Ritschel, and B.~Schiele.
\newblock What is holding back convnets for detection?
\newblock In {\em Pattern Recognition}, pages 517--528. Springer, 2015.

\bibitem{ruszczynski2006}
A.~P. Ruszczy{\'n}ski.
\newblock {\em Nonlinear optimization}, volume~13.
\newblock Princeton university press, 2006.

\bibitem{bio2}
M.~Spencer, J.~Eickholt, and J.~Cheng.
\newblock A deep learning network approach to ab initio protein secondary
  structure prediction.
\newblock {\em IEEE/ACM Trans. Comput. Biol. Bioinformatics}, 12(1):103--112,
  2015.

\bibitem{szegedy2015}
C.~Szegedy, W.~Liu, Y.~Jia, P.~Sermanet, S.~Reed, D.~Anguelov, D.~Erhan,
  V.~Vanhoucke, and A.~Rabinovich.
\newblock Going deeper with convolutions.
\newblock In {\em {IEEE} Conference on Computer Vision and Pattern Recognition
  (CVPR)}, pages 1--9, 2015.

\bibitem{szegedy2013}
C.~Szegedy, W.~Zaremba, I.~Sutskever, J.~Bruna, D.~Erhan, I.~J. Goodfellow, and
  R.~Fergus.
\newblock Intriguing properties of neural networks.
\newblock In {\em International Conference on Learning Representations (ICLR)},
  2014.

\bibitem{ostrich2015}
C.-Y. Tsai and D.~Cox.
\newblock Are deep learning algorithms easily hackable?
\newblock \url{http://coxlab.github.io/ostrichinator}.

\bibitem{vedaldi2015}
A.~Vedaldi and K.~Lenc.
\newblock Matconvnet: Convolutional neural networks for matlab.
\newblock In {\em ACM International Conference on Multimedia (MM)}, pages
  689--692, 2015.

\bibitem{walker90}
H.~F. Walker and L.~T. Watson.
\newblock Least-change secant update methods for underdetermined systems.
\newblock {\em SIAM Journal on numerical analysis}, 27(5):1227--1262, 1990.

\end{thebibliography}
}

\end{document}